\documentclass[times, review, 10pt]{elsarticle}
\usepackage[T1]{fontenc}
\usepackage{amsmath,amssymb}
\usepackage{cases}
\usepackage{xcolor}
\usepackage{graphicx}
\usepackage[normalem]{ulem}
\usepackage{subfigure}
\usepackage{caption}
\usepackage{float}
\usepackage[ruled]{algorithm2e}
\usepackage{epstopdf}
\usepackage{multirow}
\usepackage{makecell}
\usepackage{rotating}
\usepackage{booktabs}
\usepackage{tabularray}
\usepackage{setspace}
 \useunder{\uline}{\ul}{}
\usepackage[colorlinks,
linkcolor=blue,
anchorcolor=blue,
citecolor=blue,
urlcolor=magenta
]{hyperref}

\newtheorem{Theorem}{Theorem}[section]
\newtheorem{Lemma}[Theorem]{Lemma}

\usepackage{amssymb}
\usepackage{amsmath}
\usepackage{mathtools}
\journal{Pattern Recognition}

\begin{document}

\begin{frontmatter}

\title{Structured Channel Network Pruning with\\ Bi-Sparse Matrix Factorization}
\author[1,2]{Jingjing Liu} 
\ead{jjliu@shu.edu.cn} 
\author[1]{Ziyang Wang}
\ead{wangziyang@shu.edu.cn} 
\author[3]{Xianchao Xiu\corref{cor1}}
\ead{xcxiu@shu.edu.cn} 
\author[4]{Wanquan Liu}
\ead{liuwq63@mail.sysu.edu.cn} 
\affiliation[1]{organization={School of Microelectronics, Shanghai Key Laboratory of Chips and Systems for Intelligent Connected Vehicle},
            addressline={Shanghai University}, 
            city={Shanghai 200444}, 
            country={China}}
\affiliation[2]{organization={State Key Laboratory of Integrated Chips and Systems},
            addressline={Fudan University}, 
            city={Shanghai 201203}, 
            country={China}}             
\affiliation[3]{organization={School of Mechatronic Engineering and Automation},
            addressline={Shanghai University}, 
            city={Shanghai 200444}, 
            country={China}}    
\affiliation[4]{organization={School of Intelligent Systems Engineering},
            addressline={Sun Yat-sen University}, 
            city={Shenzhen 510275}, 
            country={China}}    
\cortext[cor1]{Corresponding author}       

\begin{abstract}
Structured channel pruning for convolutional neural networks (CNNs) plays an important role in pattern recognition and deep learning. Nevertheless, existing methods usually calculate channel importance scores based on the magnitudes of individual filters, making it difficult to distinguish shared redundancy from individual channel contributions. To tackle this limitation, we propose a novel structured channel pruning method termed bi-sparse matrix factorization (BSMF). Specifically, BSMF characterizes the convolutional weight matrix as output-channel coefficients over a shared basis with row-wise $\ell_{2,p}$-norm and column-wise $\ell_{2,q}$-norm regularization terms, respectively. Note that $p,q\in[0,1)$, which, to our best knowledge, has not been explored in the literature. To solve the resulting  non-convex optimization problem, we develop an efficient update scheme built upon the alternating direction method of multipliers (ADMM), in which all subproblems admit closed-form solutions. After that, a compact CNN is obtained by reconstructing the convolutional filters from the learned factors and discarding the redundant channels. Extensive experiments on the CIFAR-10 and ImageNet-1K datasets demonstrate that our proposed  BSMF achieves significant network compression while maintaining high reconstruction accuracy. The source code is publicly available at \url{https://github.com/xianchaoxiu/BSMF}.
\end{abstract}


\end{frontmatter}

\section{Introduction}
\label{sec1}
Convolutional neural networks (CNNs) have achieved considerable success in computer vision \cite{zhao2024review}. However, when deployed on mobile devices, deeper and wider networks must operate within limited memory and computation resources, which limits their practical implementation \cite{li2021survey}. In this regard, model compression techniques have been developed to reduce network complexity while preserving predictive accuracy. Interested readers are referred to recent surveys \cite{liu2025survey,khan2026model}.

Generally speaking, existing model compression methods are grouped into four categories: pruning, quantization, knowledge distillation, and low-rank decomposition. Among them, pruning eliminates redundant weights or structured units \cite{ji2025networkslimminglp}, quantization reduces the precision of weights or activations  \cite{wang2026domain}, knowledge distillation trains a compact student model under the supervision of a teacher model \cite{huang2025distplus}, and low-rank decomposition factorizes dense weight matrices or tensors into smaller components \cite{gabor2026directtensor}. Compared with the other three categories, pruning possesses the advantage of achieving flexible trade-offs between model performance and computational complexity. As discussed in \cite{cheng2024pruningsurvey}, network pruning methods fall into unstructured and structured pruning. Technically, unstructured pruning produces irregular sparsity and usually demands dedicated software or hardware support for speedup. By contrast, structured pruning discards complete convolutional filters, yielding a compact network compatible with standard dense convolution implementations \cite{yan2026csmf}. Consequently, structured pruning has garnered significant attention from both academia and industry \cite{he2024structuredsurvey}.

Within structured pruning, filters are commonly selected using  cross-layer importance evaluation \cite{lian2024crosslayer}, global channel attention \cite{wang2024globalattention}, or gradient information \cite{zhang2025gcpnet}. Other paradigms include channel masks \cite{ma2025globalmask} and dynamic strategies realized by input-dependent gates \cite{shi2024featuregate}. Nevertheless, scores based on individual filter responses may fail to characterize shared components and channel-wise contributions. Recent efforts have sought to capture interdependencies among filters using independence criteria or tensor decomposition techniques \cite{yang2024sifp,pham2024coring}. For example, efficient tensor decomposition-based filter pruning represented individual filters using higher-order singular value decomposition and identified similar filters from distances between their factors \cite{pham2024coring}. To exploit filter similarity during optimization, co-synthesizing multiple similar filters (CSMF) grouped similar filters and synchronized their updates using fused gradients and individual weight increments, consolidating their information into representative filters before removal \cite{yan2026csmf}. More recently, dynamic filter pruning via unified importance and redundancy  combined filter magnitudes with inter-filter correlations to consider filter importance and redundancy \cite{shaikh2026dfpv}. Although these methods alleviate inter-filter redundancy, they do not learn output-channel coefficients over a shared basis. Therefore, by aligning the coefficients in the decomposed representation with the output channels, one can build an explicit connection between shared structural redundancies and channel selection.

A complementary strategy is to use sparse regularization terms to induce removable channel structures. Specifically, network slimming using $\ell_p$-norm ($0<p<1$) regularization term  approximated $\ell_0$-norm sparsity \cite{ji2025networkslimminglp}, while channel pruning for convolutional neural networks using $\ell_0$-norm constraints directly constrained nonzero channel indicators \cite{chen2025l0pruner}. Although these objectives facilitate channel removal, they do not explicitly model the shared structure among filters. This motivates a  combined treatment of shared redundancy and channel sparsity. In practice, channel selection must satisfy various constraints, such as layer-width requirements, residual connections, and global resource budgets \cite{gonzalez2024dpp,ye2023pagcp}. At the deployment stage, some methods retain the decomposed factors as separate operators during inference, which can introduce additional computational overhead \cite{pham2025norton}. \textit{Therefore, a practical channel pruning framework should connect shared redundancy modeling with channel selection while preserving standard dense convolutions for deployment, without introducing additional factorized operators during inference.}

\begin{figure*}[t]
\centering
\includegraphics[width=0.98\textwidth]{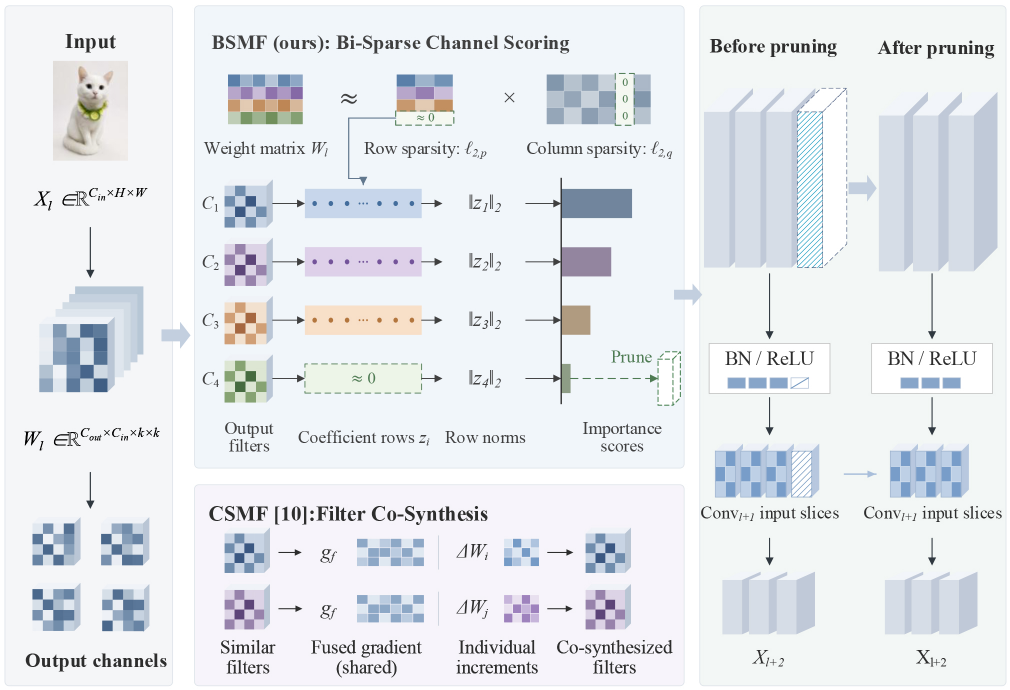}
\caption{Comparison between our BSMF and CSMF \cite{yan2026csmf}, where BSMF scores channels using the row-wise norms of $Z_A$, while CSMF co-synthesizes filters using fused gradients and individual weight increments.}
\label{fig:bsmf_framework}
\end{figure*}

Motivated by the above observations, we propose bi-sparse matrix factorization (BSMF) for structured channel pruning.
This method decomposes each convolutional weight matrix into channel coefficients and a shared basis, where each output channel corresponds to a specific coefficient vector. In addition, we impose structured sparsity on the rows of the coefficient factor and the columns of the basis factor. Here, the row-wise norms of the coefficient are used as a scoring criterion to select channels, while adhering to constraints on layer width, connectivity, and computational resources. Before removing the channels, the factors are recombined to reconstruct the convolutional filters, ultimately yielding a compact network composed of standard dense convolutions. Fig. \ref{fig:bsmf_framework} illustrates the differences between our proposed BSMF and existing CSMF \cite{yan2026csmf}. It can be seen that CSMF coordinates the update process of filters by combining a common fusion gradient with individual weight increments, while our BSMF constructs channel scores by utilizing shared convolutional filter information, and each output filter is represented by a row of coefficients over the shared basis. Furthermore, the norm of this row determines the priority for retaining the corresponding filter, which provides greater flexibility.

The main contributions of this paper can be summarized as follows.
\begin{itemize}
\item We propose a novel matrix factorization model by imposing non-convex structured regularization on both the output-channel coefficients and the shared basis. By leveraging the row-wise norms of the learned coefficients, we derive channel importance scores that account for filter relationships, thereby providing guidance for structured channel pruning.

\item We develop an efficient optimization algorithm based on the alternating minimization framework, where both variables can be solved using ADMM. Although establishing strict theoretical convergence is challenging, we empirically verify the convergence numerically.

\item We demonstrate the superiority of the proposed method through sufficient experiments. Moreover, we visualize the learned importance scores using channel-selection heatmaps and quantify the selection differences by comparing the sets of retained channels via the Jaccard coefficient.
\end{itemize}

The remainder of this paper is organized as follows.
Section \ref{sec:related_work} reviews the related work.
Section \ref{sec:methodology} presents our mathematical formulation, optimization algorithm, and pruning procedure.
Section \ref{sec:experiments} reports the experimental setup, numerical results, ablation studies, and further discussions.
Section \ref{sec:conclusion} concludes the paper.

\section{Related Work}\label{sec:related_work}

This section reviews recent work on channel pruning, structured pruning, and sparse regularization.

\subsection{Channel Pruning}
Channel scoring must distinguish information content from inter-channel redundancy. In previous studies, filter pruning using high-rank feature map  ranks feature responses but may overlook cross-channel duplication \cite{lin2020hrank}. Later, channel independence-based pruning \cite{sui2021chip} and automatic network pruning via Hilbert-Schmidt independence criterion lasso under information bottleneck principle \cite{guo2023apib} capture channel dependence or information preservation. However, these methods rely on activation statistics derived from sampling when assigning scores. Although the above methods reduce redundancy, they do not learn sparse channel coefficients within a shared-basis framework. Beyond similarity-based pruning, more recent methods including singular value-driven automated filter pruning \cite{pham2025singularpruning}, adding before pruning (ABP) \cite{tian2025abp}, and iterative pruning with prompt knowledge distillation (IPPKD) \cite{li2026ippkd} preserve information through spectral structure, attention-guided fusion, or distillation, respectively. By contrast, our BSMF uses the row-wise norms of the shared-basis coefficients to score channels, aligning scoring with sparse optimization yet without activation statistics.

\subsection{Structured Pruning}
Structured pruning removes complete filters, and existing methods mainly rely on masks, filter similarity, or factorized representations. Although generative adversarial learning (GAL) \cite{lin2019gal} and leanflex grouped filter pruning (LeanFlex-GKP) \cite{zhang2025leanflexgkp} learn structural masks or group configurations, they do not explicitly represent shared filter components. Besides, tensor decomposition-based filter pruning \cite{pham2024coring} and CSMF \cite{yan2026csmf} characterize inter-filter similarity through decomposition-factor distances and coordinated fusion, respectively.
However, these methods do not learn channel-aligned sparse coefficients over a shared basis  \cite{pham2025norton}.
Our BSMF addresses these limitations by learning a shared representation for channel selection and recombining its factors before removal, thereby preserving standard dense convolutions at deployment.

\subsection{Sparse Regularization}
Sparse regularization methods also differ in the granularity of the structures they sparsify. Beyond explicit channel regularizations, feature flow regularization  promotes structured sparsity through feature propagation \cite{wu2023featureflow}. More recently, double sparse factorization  learns two element-wise sparse factor matrices \cite{boza2025double}. Nevertheless, feature-level or element-wise sparsity does not necessarily eliminate an entire output-channel group. In contrast, our BSMF couples weight reconstruction with $\ell_{2,p}$-norm regularization on coefficient rows and  $\ell_{2,q}$-norm regularization on basis columns. The row regularization suppresses complete output-channel groups, whereas the basis-column regularization encourages a compact shared representation.

\section{Methodology}\label{sec:methodology}

This section first introduces the proposed formulation and optimization algorithm, followed by the pruning strategy.

\subsection{Our Formulation}

For a convolutional layer, unfold its weights into $W\in\mathbb{R}^{m\times n}$ with $m=C_{\mathrm{out}}$ and
$n=C_{\mathrm{in}}h^2$, where $C_{\mathrm{out}}$ and $C_{\mathrm{in}}$ denote the number of output and input channels, respectively, and $h$ represents the width of the convolution filter. For the sake of convenience, we omit the layer index and approximate the weight matrix as
\begin{equation}
    W \approx AB,
    \label{eq:ab_decomposition}
\end{equation}
where $A\in\mathbb{R}^{m\times r}$ and $B\in\mathbb{R}^{r\times n}$,
with integer rank $1\leq r \leq \min(m,n)$. In fact, each row of  $A$ corresponds to an output channel, while the rows of  $B$ form a shared basis and its columns correspond to the input-filter coordinates.

To enhance pruning capability, we consider the following model
\begin{equation}
\min_{A,B}\quad \frac{1}{2}\|W-AB\|_{F}^{2} +\lambda_A\|A\|_{2,p}^{p} +\lambda_B\|B\|_{2,q}^{q},
\label{eq:main_obj}
\end{equation}
where $\|\cdot\|_F$ is the Frobenius norm,
$\lambda_A,\lambda_B\geq0$ are parameters to weight the two sparsity-inducing regularization terms. We would like to point out that
\begin{equation}
\|A\|_{2,p}^{p}=\sum_{i=1}^{m}\|A_{i,:}\|_2^p,\quad \|B\|_{2,q}^{q}=\sum_{j=1}^{n}\|B_{:,j}\|_2^q
\label{eq:main_obj}
\end{equation}
where $A_{i,:}$ denotes the $i$-row of $A$ and $B_{:,j}$ denotes the $j$-th column of $B$. This means that $\|A\|_{2,p}^{p}, \|B\|_{2,q}^{q}$ are row-wise and column-wise structured sparsity, respectively. In addition, $p, q \in [0, 1)$, which includes $p, q \in (0, 1)$ in \cite{li2023sparse} and the special case $p, q = 0$ in \cite{liu2027blufs}. Specifically, when $p, q = 0$, they are defined as the counts of nonzero rows and nonzero columns, respectively.

\subsection{Optimization Algorithm}

Problem \eqref{eq:main_obj} involves two variables, making direct optimization quite difficult. Inspired by \cite{boza2025double}, we adopt an alternating minimization scheme by computing
\begin{equation}
\begin{cases}
A^{k+1}=\arg\min_A ~\displaystyle\frac{1}{2}\|W-AB^k\|_{F}^{2} +\lambda_A \|A\|_{2,p}^{p},\\
B^{k+1}=\arg\min_B ~\displaystyle\frac{1}{2}\|W-A^{k+1}B\|_{F}^{2} +\lambda_B \|B\|_{2,q}^{q}.
\end{cases}
\end{equation}
In what follows, we describe how to solve the subproblems using the alternating direction method of multipliers (ADMM).

\subsubsection{$A$-subproblem}
With $B^k$ fixed, $A$-subproblem can be reformulated as
\begin{equation}
\begin{aligned}
\min_{A, Z_A}
\quad &  \frac{1}{2}\|W-AB^k\|_F^2
+ \lambda_A\|Z_A\|_{2,p}^p \\
\mathrm{s.t.}
\quad&  A=Z_A.
\end{aligned}
\label{eq:a_sparse_problem}
\end{equation}
The corresponding augmented Lagrangian is
\begin{equation}
\mathcal{L}_{A}(A,Z_A;U_A) =
\frac{1}{2}\|W-A B^k\|_F^2
+\lambda_A\|Z_A\|_{2,p}^p +\langle U_A,A-Z_A\rangle
+\frac{\rho}{2}\|A-Z_A\|_F^2,
\label{eq:a_aug_lag}
\end{equation}
where $\langle\cdot,\cdot\rangle$ denotes the inner product and $\rho>0$ is the tuning parameter.

\begin{itemize}
    \item Update $A$ according to
\begin{equation}
A^{k+1}
=
\arg\min_A ~\frac{1}{2}\|W-A B^k\|_{F}^{2}
+\frac{\rho}{2} \|A-Z_A^k+ U_A^k/\rho\|_{F}^{2}.
\label{eq:a_subproblem}
\end{equation}
Setting the gradient with respect to $A$ to zero yields
\begin{equation}
(A^{k+1} B^k-W)B^{k,\top} +\rho(A^{k+1}-Z_A^k+U_A^k/\rho)=0,
\end{equation}
where $B^{k,\top}$ denotes the transpose of $B^{k}$.
Therefore, the closed-form solution of problem \eqref{eq:a_subproblem} is 
\begin{equation}
A^{k+1}
=
(WB^{k,\top}+\rho Z_A^k-U_A^k)
(B^k B^{k,\top} + \rho I)^{-1},
\label{eq:a_update}
\end{equation}
where $I\in\mathbb{R}^{r\times r}$ is the identity matrix.
    \item Update $Z_A$ by 
    \begin{equation}
    \begin{aligned}
 Z_A^{k+1}
&=
\arg\min_{Z_A}~
\lambda_A \|Z_A\|_{2,p}^{p} +\frac{\rho}{2}
\|A^{k+1}-Z_A+U_A^k/\rho\|_{F}^{2}\\ 
&=
\arg\min_{Z_A} ~\lambda_A \|Z_A\|_{2,p}^{p}
+\frac{\rho}{2}\|Z_A-V_A^{k+1}\|_{F}^{2},
    \end{aligned}
\label{eq:za_subproblem}
\end{equation}
where $V_A^{k+1}=A^{k+1}+U_A^k/\rho$. Thus, the closed-form is
\begin{equation}
Z_A^{k+1}
\in
\operatorname{prox}_{\frac{\lambda_A}{\rho}\|\cdot\|_{2,p}^{p}}
(V_A^{k+1}).
\label{eq:za_prox}
\end{equation}
Specifically, for the $i$-th row, it becomes
\begin{equation}
(Z_A^{k+1})_{i,:}
\in
\begin{cases}
\operatorname{prox}_{\frac{\lambda_A}{\rho}|\cdot|^p}(\|(V_A^{k+1})_{i,:}\|_2)
\cdot\frac{(V_A^{k+1})_{i,:}}{\|(V_A^{k+1})_{i,:}\|_2}, & \|(V_A^{k+1})_{i,:}\|_2>0,\\ 
\{0\}, & \|(V_A^{k+1})_{i,:}\|_2=0,
\end{cases}
\label{eq:za_group_update}
\end{equation}
and  $\operatorname{prox}$ is defined in the following lemma.
\begin{Lemma}\label{P}
For $x\in \mathbb{R}$ and $p\in [0,1)$, consider the proximal operator 
\begin{equation}
\begin{aligned} \label {U4}
    \operatorname{prox}_{\lambda |\cdot|^{p}}(t) & =\arg\min_x~\lambda|x|^p+\frac{1}{2}(x-t)^2\\
    & = \begin{cases}\{0\}, & |t|<\kappa(\lambda, p), \\
    \{0, \operatorname{sgn}(t) c(\lambda, p)\}, & |t|=\kappa(\lambda, p), \\
    \{\operatorname{sgn}(t) \varpi_p(|t|)\}, & |t|>\kappa(\lambda, p),\end{cases}
\end{aligned}
\end{equation}
where
\begin{equation}
\begin{aligned}
c(\lambda, p)&=(2 \lambda(1-p))^{\frac{1}{2-p}}>0,\\
\kappa(\lambda, p)&=(2-p) \lambda^{\frac{1}{2-p}}(2(1-p))^{\frac{p+1}{p-2}},\\
\varpi_p(t) &\in \{x \mid x-t+\lambda p \operatorname{sgn}(x) x^{p-1}=0, x>0 \}.
\end{aligned}
\end{equation}
More details can be found in \cite{xiu2025bi}.
\end{Lemma}
    \item Update $U_A$ via
\begin{equation}
    U_A^{k+1}
    =
    U_A^k
    +
    \rho (
    A^{k+1}
    -
    Z_A^{k+1}).
    \label{eq:ua_update}
\end{equation}
\end{itemize}

\subsubsection{$B$-subproblem}
With the updated $Z_A^{k+1}$ fixed, $B$-subproblem can be written as
\begin{equation}
\begin{aligned}
\min_{B,Z_B} \quad &  \frac{1}{2}\|W-Z_A^{k+1}B\|_F^2 + \lambda_B\|Z_B\|_{2,q}^q \\
\mathrm{s.t.} \quad & B=Z_B,
\end{aligned}
\label{eq:b_sparse_problem}
\end{equation}
and the augmented Lagrangian is
\begin{equation}
\begin{aligned}
\mathcal{L}_{B}(B,Z_B;U_B) =\frac{1}{2}\|W-Z_A^{k+1}B\|_F^2
+\lambda_B\|Z_B\|_{2,q}^q +\langle U_B,B-Z_B\rangle
+\frac{\rho}{2}\|B-Z_B\|_F^2.
\end{aligned}
\label{eq:b_aug_lag}
\end{equation}
Similar to the above updates, the following iterative forms can be obtained.
\begin{itemize}
    \item Update $B$ by
  \begin{equation}
B^{k+1}
=
(Z_A^{{k+1},\top} Z_A^{k+1}+\rho I)^{-1}
(Z_A^{{k+1},\top} W+\rho Z_B^k-U_B^k),
\label{eq:b_update}
\end{equation}  
where where $Z_A^{{k+1},\top}$ denotes the transpose of $Z_A^{{k+1}}$.
    \item Update $Z_B$ by 
    \begin{equation}
Z_B^{k+1}
\in
\operatorname{prox}_{\frac{\lambda_B}{\rho}\|\cdot\|_{2,q}^{q}}
(V_B^{k+1}).
\label{eq:zb_prox}
\end{equation}
where $V_B^{k+1}=B^{k+1}+U_B^k/\rho$ and its column is defined as
\begin{equation}
(Z_B^{k+1})_{:,j}
\in
\begin{cases}
\operatorname{prox}_{\frac{\lambda_B}{\rho}|\cdot|^q}(\|(V_B^{k+1})_{:,j}\|_2)
\cdot\frac{(V_B^{k+1})_{:,j}}{\|(V_B^{k+1})_{:,j}\|_2}, & \|(V_B^{k+1})_{:,j}\|_2>0,\\ 
\{0\}, & \|(V_B^{k+1})_{:,j}\|_2=0.
\end{cases}
\label{eq:zb_group_update}
\end{equation}
    \item Update $U_B$ by
    \begin{equation}
U_B^{k+1}
=
U_B^k
+
\rho (
B^{k+1}
-
Z_B^{k+1}
 ).
\label{eq:ub_update}
\end{equation}
\end{itemize}

\subsubsection{Overall Scheme}
After $K$ outer iterations, the weight matrix is reconstructed as
\begin{equation}
    \hat{W}
    =
    Z_A^K Z_B^K.
    \label{eq:w_bsmf}
\end{equation}
Finally, Algorithm \ref{alg:fgsp_admm} summarizes the initialization and alternating updates.

\begin{algorithm}[t]
\setstretch{1}
\SetAlgoLined
\DontPrintSemicolon
\SetKwSty{textbf}
\SetAlgoCaptionSeparator{ }
\SetKwInput{Input}{Input}
\SetKwInput{Output}{Output}
\SetKwBlock{LeftADMM}{$A$-subproblem}{end}
\SetKwBlock{RightADMM}{$B$-subproblem}{end}
\caption{Alternating Minimization for Solving Problem \eqref{eq:main_obj}}\label{alg:fgsp_admm}
\Input{Weight $W\in\mathbb{R}^{m\times n}$, rank $r$, 
iteration $K$, parameters $\lambda_A,\lambda_B,p,q,\rho$.}
{\SetAlgoHangIndent{0pt}
\textbf{Initialize:} Compute $W_r=P_r\Sigma_r Q_r^\top$, let
$Z_A^0=P_r\Sigma_r^{1/2}$, $Z_B^0=\Sigma_r^{1/2}Q_r^\top$, $U_A^0,U_B^0=0$}.\;
\While{$k<K$}{
    \LeftADMM{
        Update $A^{k+1}$ using Eq. \eqref{eq:a_update}.\;
        Update $Z_A^{k+1}$ using Eq. \eqref{eq:za_prox}.\;
        Update $U_A^{k+1}$ using Eq. \eqref{eq:ua_update}.\;
    }
    \RightADMM{
        Update $B^{k+1}$ using Eq. \eqref{eq:b_update}.\;
        Update $Z_B^{k+1}$ using Eq. \eqref{eq:zb_prox}.\;
        Update $U_B^{k+1}$ using Eq. \eqref{eq:ub_update}.\;
    }
    $k\leftarrow k+1$.\;
}
\Output{$Z_A^K,Z_B^K$ and 
compute $\hat{W}$ using Eq. \eqref{eq:w_bsmf}.}
\end{algorithm}

\subsection{Pruning Strategy}
\label{sec:channel_selection}

This part describes how to use the factor matrix $Z_{A}^{K}$ to rank output channels, construct masks, and then convert the masked model into a compact dense network.

\subsubsection{Channel Importance}

Let $C_l=m_l$ denote the number of output channels in the $l$-th layer. Its importance score is defined as
\begin{equation}
s_{l,i}=\|Z_{A,l}^{K}(i,:)\|_2.
\label{eq:pipeline_channel_score}
\end{equation}
A smaller score indicates a weaker coefficient group and a lower retention priority. Since the raw score scales may vary across different layers, the scores are normalized independently for each layer. Let $a_l$ and $b_l$ be the 5th and 95th percentiles of the scores in the $l$-th layer, respectively. When $b_l-a_l\geq\epsilon$, define
\begin{equation}
\widetilde s_{l,i}=\operatorname{clip}_{[0,1]} \left(
\frac{s_{l,i}-a_l}{b_l-a_l+\epsilon}\right),
\label{eq:pipeline_score_normalization}
\end{equation}
where $\epsilon=10^{-12}$ and $\operatorname{clip}_{[0,1]}(x)=\min\{1,\max\{0,x\}\}$. This robust scaling limits the effect of extreme scores while placing layers on a common range. If $b_l-a_l<\epsilon$, all normalized scores for that layer are set to one, thereby avoiding division by an extremely small value.

\subsubsection{Mask Selection}
Let $L_{\mathrm{prune}}\subseteq L$ be the set of layers eligible for output channel pruning, and let
$N=\sum_{l\in L_{\mathrm{prune}}}C_l$ be the total number of channels across these layers. For a given nominal pruning ratio $\pi \in [0, 1)$, the normalized scores of all eligible channels are sorted in ascending order as $t_{(1)} \leq \cdots \leq t_{(N)}$. In addition, the initial threshold is defined as the following empirical lower-tail order statistic
\begin{equation}
j_{\pi}=\min \left\{\lfloor\pi N\rfloor+1,N\right\},\quad
\tau_{\mathrm{ch}}=t_{(j_{\pi})}.
\end{equation}
The initial mask is defined as
\begin{equation}
m_{l,i}^{(0)}=
\begin{cases}
1,&\widetilde s_{l,i}>\tau_{\mathrm{ch}},\\
0,&\widetilde s_{l,i}\leq\tau_{\mathrm{ch}},
\end{cases}
\label{eq:pipeline_global_threshold}
\end{equation}
where $1$ indicates that the channel is retained, and $0$ indicates that it is marked for removal. When all channel scores are distinct, this threshold results in the removal of the $j_{\pi}$ channels with the lowest scores, and if any channels have scores equal to the threshold, they are also removed. Thus, $\pi$ determines the nominal lower-tail proportion.

Subsequently, the threshold mask must be refined to satisfy constraints regarding layer width and connectivity. Let $c_l^-$ and $c_l^+$ be the allowed minimum and maximum retained widths, where $1\leq c_l^-\leq c_l^+\leq C_l$. If a layer retains fewer than $c_l^-$ channels, its highest-scoring discarded
channels are restored until the lower bound is met. If it retains more than $c_l^+$, its lowest-scoring retained channels are removed until the upper bound is met. Let $m_{l,i}$ denote the resulting mask and $\mathcal I_l=\{i:m_{l,i}=1\}$ be the corresponding retained index set. At the adjustment boundaries, if multiple indices share the same score, it is possible to derive several valid masks. Furthermore, if removing certain channels would violate the width constraints imposed by residual addition operations, those channels are protected and retained.

\subsubsection{Network Construction}

For each layer, the dense convolutional filter is reconstructed using Eq. \eqref{eq:w_bsmf}, and the output filters indexed by $\mathcal I_l$ are retained. Simultaneously, the associated normalization parameters and input-channel slices are removed from the subsequent connected layers. After that, the batch-normalization statistics are recalibrated, and the compact model is fine-tuned to recover its accuracy.

\section{Experiments}\label{sec:experiments}

To validate the superiority of our proposed BSMF, some state-of-the-art pruning methods are selected, including HRank\footnote{\url{https://github.com/lmbxmu/HRank}}, CHIP\footnote{\url{https://github.com/Eclipsess/CHIP_NeurIPS2021}}, APIB\footnote{\url{https://github.com/sunggo/APIB}}, CORING\footnote{\url{https://github.com/vantienpham/CORING}}, NORTON\footnote{\url{https://github.com/vantienpham/NORTON}}, CSMF\footnote{\url{https://github.com/row-niu/CSMF}}, DFPV\footnote{\url{https://github.com/MZhangXin/dfpv}}, GAL\footnote{\url{https://github.com/anonymouscvpr1983/GAL}}, LeanFlex-GKP\footnote{\url{https://github.com/MorrisZJ/LeanFlex-GKP}}, $\ell_0$\footnote{\url{https://github.com/TCCofWANG/l0_prune}}  $\ell_p$\footnote{\url{https://github.com/Fengrui-Ji/network_pruning}}, ABP, and IPPKD\footnote{\url{https://github.com/tlyslll/IPPKD}}.

Subsection \ref{Datasets} describes the datasets, models, and training settings. Subsection \ref{Comparison} evaluates the proposed BSMF on the CIFAR-10 and ImageNet-1K datasets in terms of accuracy, floating-point operations (FLOPs), and the number of parameters. Subsection \ref{Ablation} examines the roles of individual components, including matrix factorization, bi-sparsity optimization, and channel scoring. Subsection \ref{Discussion} provides further analysis of the convergence, resulting channel selections, sensitivity to the regularization order, and distribution of effective weights.

\subsection{Experimental Setup}\label{Datasets}

\subsubsection{Datasets and Models}
Experiments are conducted on the CIFAR-10\footnote{\url{https://www.cs.toronto.edu/~kriz/cifar.html}} dataset using ResNet-56, ResNet-110, and VGG-16 models, while the ResNet-50 model is used for the ImageNet-1K\footnote{\url{https://www.image-net.org/challenges/LSVRC/2012/}} dataset. Here, the CIFAR-10 dataset  consists of 50,000 training images and 10,000 testing images across 10 classes, with an image size of $32 \times 32$ pixels. The ImageNet-1K dataset contains approximately 1.28 million training images and 50,000 validation images across 1,000 classes.

\subsubsection{Training Settings}

During the fine-tuning stage, stochastic gradient descent (SGD) is employed. For the CIFAR-10 dataset, the batch size is set to 128 for the ResNet-56, ResNet-110, and VGG-16 models. The learning rates for ResNet-56 at 40\% FLOPs reduction, ResNet-110 at approximately 50\% FLOPs reduction, and VGG-16 are 0.009, 0.01, and 0.015, respectively. The corresponding momentum values are 0.95, 0.95, and 0.97, with weight decay $5\times10^{-4}$ for all three. For the ImageNet-1K dataset, ResNet-50 is fine-tuned using Nesterov SGD with a batch size of 256, a learning rate of 0.025, a momentum of 0.9, and a weight decay coefficient of $5\times10^{-5}$. Furthermore, normalization parameters and bias terms are excluded from weight decay. For ResNet-56 and ResNet-110, all comparison methods are reimplemented with their recommended hyperparameters, using the same pretrained checkpoint, pruning targets, and fine-tuning protocol within each architecture. For VGG-16 and ResNet-50, the comparison results are taken directly from the corresponding papers, together with their original training recipes and dense baselines.

\subsubsection{Evaluation Metrics}

To evaluate the performance of the comparative methods, Top-1 and Top-5 accuracy are used, representing the proportion of cases where the ground-truth label appears within the top 1 and top 5 categories, respectively, based on the predicted scores \cite{russakovsky2015imagenet}. A higher value indicates better performance. Additionally, FLOPs and the number of parameters (Params) \cite{lin2020hrank} are considered, where lower values are preferred when accuracy is comparable. For any of the aforementioned resource metrics, the pruning ratio (PR) is calculated as 
\begin{equation}
\mathrm{PR} =
\left(1-\frac{C_{\mathrm{pruned}}}{C_{\mathrm{dense}}}\right)
\times 100\%.
\end{equation}
where $C_{\mathrm{dense}}$ and $C_{\mathrm{pruned}}$ are the corresponding
counts for the dense baseline and pruned model. A higher PR value indicates a greater relative reduction.

\begin{table*}[t]
\centering
\renewcommand\arraystretch{1.2}

\caption{Pruning results for ResNet-56 and ResNet-110 on the CIFAR-10 dataset, where the best and second-best pruned values are shown in bold
and underlined, respectively.}
\label{CIFAR-10}
\vskip-0.2cm
\setlength{\tabcolsep}{4.5 pt}

\fontsize{7pt}{7pt}\selectfont
\begin{tabular*}{\textwidth}{@{\extracolsep{\fill}}llccc}
\toprule
\textbf{Model} & \textbf{Method} & \textbf{Top-1 Accuracy} & \textbf{FLOPs (PR)} & \textbf{Params (PR)} \\
\midrule
\multirow{9}{*}{\textbf{ResNet-56}} & Baseline & 93.26\% & 127.09M (0.0\%) & 0.85M (0.0\%) \\
& $\ell_p$ \cite{ji2025networkslimminglp} & 93.87\% & \textbf{75.62M (40\%)} & 0.56M (34\%) \\
& CORING \cite{pham2024coring} & 93.79\% & 76.12M (40\%) & \underline{0.47M (45\%)} \\
& APIB \cite{guo2023apib} & \underline{93.92\%} & 76.35M (40\%) & 0.53M (37\%) \\
& \textbf{BSMF (Ours)} & \textbf{94.18\%} & \underline{75.92M (40\%)} & \textbf{0.44M (48\%)} \\
\cmidrule(lr){2-5}
& HRank \cite{lin2020hrank} & 92.49\% & \underline{62.72M (50\%)} & 0.49M (42\%) \\
& NORTON \cite{pham2025norton} & 92.41\% & 63.33M (50\%) & 0.38M (55\%) \\
& $\ell_0$ Pruner \cite{chen2025l0pruner} & \underline{93.03\%} & 64.04M (50\%) & \underline{0.36M (58\%)} \\
& \textbf{BSMF (Ours)} & \textbf{93.84\%} & \textbf{62.23M (50\%)} & \textbf{0.34M (60\%)} \\
\midrule
\multirow{8}{*}{\textbf{ResNet-110}} & Baseline & 93.50\% & 257.09M (0.0\%) & 1.73M (0.0\%) \\
& $\ell_p$ \cite{ji2025networkslimminglp} & 94.41\% & \underline{126.70M (50\%)} & 1.11M (36\%) \\
& APIB \cite{guo2023apib} & \underline{94.45\%} & 127.29M (50\%) & \textbf{0.88M (49\%)} \\
& \textbf{BSMF (Ours)} & \textbf{94.46\%} & \textbf{125.00M (51\%)} & \underline{0.89M (48\%)} \\
\cmidrule(lr){2-5}
& $\ell_0$ Pruner \cite{chen2025l0pruner} & 93.26\% & 114.92M (55\%) & \underline{0.69M (60\%)} \\
& HRank \cite{lin2020hrank} & 92.96\% & 105.63M (58\%) & 0.72M (60\%) \\
& NORTON \cite{pham2025norton} & \underline{93.58\%} & \underline{102.84M (60\%)} & \textbf{0.65M (62\%)} \\
& \textbf{BSMF (Ours)} & \textbf{94.05\%} & \textbf{102.04M (60\%)} & \underline{0.69M (60\%)} \\
\bottomrule
\end{tabular*}
\end{table*}

\subsection{Comparison Results}\label{Comparison}

\begin{table*}[t]
\centering
\renewcommand\arraystretch{1.2}

\caption{Pruning results for VGG-16 on the CIFAR-10 dataset, where the best and second-best
pruned values are shown in bold and underlined, respectively.}
\label{tab:vgg16_cifar10}
\vskip-0.2cm
\setlength{\tabcolsep}{4.5 pt}

\fontsize{7pt}{7pt}\selectfont
\begin{tabular*}{\textwidth}{@{\extracolsep{\fill}}llccc}
\toprule
\textbf{Model} & \textbf{Method} & \textbf{Top-1 Accuracy} & \textbf{FLOPs (PR)} &
\textbf{Params (PR)} \\
\midrule
\multirow{11}{*}{\textbf{VGG-16}} & Baseline & 93.96\% & 313.73M (0\%) & 14.98M (0\%) \\
& LeanFlex-GKP \cite{zhang2025leanflexgkp} & 94.15\% & 178.17M (43\%) & 8.29M (44\%) \\
& GAL-0.1 \cite{lin2019gal} & 93.42\% & 171.89M (45\%) & 2.67M (82\%) \\
& HRank \cite{lin2020hrank} & 93.43\% & 145.61M (54\%) & \textbf{2.51M (83\%)} \\
& DFPV \cite{shaikh2026dfpv} & 93.76\% & 139.30M (56\%) & 3.25M (78\%) \\
& IPPKD \cite{li2026ippkd} & 94.24\% & 139.36M (56\%) & 6.89M (54\%) \\
& CHIP \cite{sui2021chip} & 93.86\% & 131.17M (58\%) & 2.76M (82\%) \\
& CORING \cite{pham2024coring} & \underline{94.42\%} & 131.44M (58\%) & 2.76M (82\%) \\
& APIB \cite{guo2023apib} & 94.08\% & 125.48M (60\%) & 3.60M (76\%) \\
& $\ell_p$ \cite{ji2025networkslimminglp} & 94.36\% & \underline{123.60M (61\%)} & 2.67M (82\%) \\
& \textbf{BSMF (Ours)} & \textbf{94.48\%} & \textbf{122.98M (61\%)} & \underline{2.66M (82\%)} \\
\bottomrule
\end{tabular*}
\end{table*}

Table \ref{CIFAR-10} shows that our BSMF achieves the highest accuracy across the evaluated pruning settings. Its gain over the strongest competing method increases from 0.26\% and 0.01\% under mild pruning to 0.63\% and 0.47\% under tight pruning on ResNet-56 and ResNet-110, respectively. Its shared-representation-based channel scoring captures inter-filter dependencies, helping BSMF retain informative and complementary channels under tight pruning budgets, thereby retaining a more informative channel subset under the same pruning budget. Under mild pruning, ranking differences have limited impact because many redundant channels remain. Under tight pruning, however, an incorrect ranking causes a larger loss of feature coverage, making channel-selection quality decisive. The ablation results in Subsection \ref{sec:ablation} support this interpretation. Overall, our BSMF achieves a more favorable accuracy-efficiency trade-off under comparable FLOPs and parameter constraints.

Table \ref{tab:vgg16_cifar10} reports the pruning results of VGG-16 on the CIFAR-10 dataset. Under the high-compression setting, our BSMF achieves an accuracy of 94.48\%, the highest among the compared methods, while substantially reducing both computation and model size. This result shows that BSMF preserves classification performance even when pruning is aggressive. Together with the results on ResNet-56 and ResNet-110, it demonstrates that our BSMF is effective on the CIFAR-10 dataset across both sequential architectures such as VGG-16 and residual architectures, rather than being limited to a particular network design.

\begin{table*}[t]
\centering
\renewcommand\arraystretch{1.2}

\caption{Pruning results for ResNet-50 on the ImageNet-1K dataset, where the best and
second-best pruned values are shown in bold and underlined, respectively.}
\label{ImageNet-1K}
\vskip-0.2cm
\setlength{\tabcolsep}{4.5 pt}

\fontsize{7pt}{7pt}\selectfont
\begin{tabular*}{\textwidth}{@{\extracolsep{\fill}}llcccc}
\toprule
\textbf{Model} & \textbf{Method} & \textbf{Top-1 Accuracy} & \textbf{Top-5 Accuracy} & \textbf{FLOPs (PR)} & \textbf{Params (PR)} \\
\midrule
\multirow{11}{*}{\textbf{ResNet-50}} & Baseline \cite{he2016resnet} & 76.15\% & 92.87\% & 4.12B (0.0\%) & 25.50M (0.0\%) \\
& LeanFlex-GKP \cite{zhang2025leanflexgkp} & 75.62\% & N/A & 2.76B (33\%) & 17.80M (30\%) \\
& GAL-0.5 \cite{lin2019gal} & 71.95\% & 90.94\% & 2.33B (43\%) & 21.20M (17\%) \\
& ABP \cite{tian2025abp} & 74.80\% & 92.36\% & 2.36B (43\%) & \textbf{13.01M (49\%)} \\
& HRank-1 \cite{lin2020hrank} & 74.98\% & 92.33\% & 2.30B (44\%) & 16.15M (37\%) \\
& $\ell_0$ Pruner \cite{chen2025l0pruner} & 75.18\% & 92.39\% & 2.27B (45\%) & 13.39M (48\%) \\
& $\ell_p$ \cite{ji2025networkslimminglp} & 76.06\% & \underline{92.91\%} & 2.22B (46\%) & 17.14M (33\%) \\
& CSMF \cite{yan2026csmf} & 75.81\% & N/A & 2.20B (47\%) & 14.92M (42\%) \\
& CHIP \cite{sui2021chip} & \underline{76.15\%} & \underline{92.91\%} & 2.11B (49\%) & 14.23M (44\%) \\
& DFPV \cite{shaikh2026dfpv} & 75.89\% & \textbf{93.04\%} & \textbf{1.99B (52\%)} & 14.05M (45\%) \\
& \textbf{BSMF (Ours)} & \textbf{76.27\%} & 92.89\% & \underline{2.07B (50\%)} & \underline{13.03M (49\%)} \\
\bottomrule
\end{tabular*}
\end{table*}

Furthermore, Table \ref{ImageNet-1K} presents the pruning results for ResNet-50 on the ImageNet-1K dataset. Our BSMF achieves the highest reported Top-1 accuracy of 76.27\%, exceeding CHIP by 0.12\% while requiring fewer FLOPs (2.07B versus 2.11B) and parameters (13.03M versus 14.23M). This performance is consistent with the coupling of reconstruction and sparse channel selection in BSMF. The reconstruction term penalizes deviations from the original weights, while row-wise regularization suppresses weak channel coefficients in the shared basis. Ranking channels by these regularized coefficient norms carries the learned suppression directly into pruning decisions, aligning channel selection with sparse optimization. Together with the CIFAR-10 results, this finding supports the effectiveness of BSMF across the evaluated architectures and its applicability to large-scale image classification.

Overall, the results on the CIFAR-10 and ImageNet-1K datasets demonstrate that our BSMF achieves a favorable accuracy-efficiency trade-off across both sequential and residual architectures. The larger accuracy gains under more aggressive pruning budgets on ResNet-56 and ResNet-110 further show its effectiveness when the retained model capacity is limited.

\begin{table*}[t]
\centering
\renewcommand\arraystretch{1.2}

\caption{Component ablation results for ResNet-56 on the CIFAR-10 dataset, where the best and
second-best pruned values are shown in bold and underlined, respectively.}
\label{tab:component_ablation}
\vskip-0.2cm
\setlength{\tabcolsep}{4.5 pt}

\fontsize{7pt}{7pt}\selectfont
\begin{tabular*}{\textwidth}{@{\extracolsep{\fill}}lcclccc}
\toprule
\textbf{Method} & \textbf{Factor.} & \textbf{Sparse opt.} &
\textbf{Channel score} & \textbf{Top-1 Accuracy (\%)} &
\textbf{FLOPs Red.} & \textbf{Params} \\
\midrule
Baseline & No & No & None & 93.26 & 0\% & 0.85M \\
BSMF without pruning & Yes & Yes & None & $93.09\pm0.02$ & 0\% & 0.85M \\
$\ell_2$ pruning & No & No & filter $\ell_2$-norm & $93.65\pm0.23$ & 40\% & 0.55M \\
BSMF + filter $\ell_2$ & Yes & Yes & filter $\ell_2$-norm & $93.74\pm0.10$ & 40\% & 0.52M \\
BSMF + filter $\ell_1$ & Yes & Yes & filter $\ell_1$-norm & $\underline{93.91\pm0.09}$ & 40\% & \underline{0.51M} \\
Factorization-only pruning & Yes & No & $A$-row norm & $93.83\pm0.18$ & 40\% & \textbf{0.44M} \\
\textbf{BSMF (Ours)} & \textbf{Yes} & \textbf{Yes} & \textbf{$A$-row norm} & $\boldsymbol{93.99\pm0.07}$ & 40\% & \textbf{0.44M} \\
\bottomrule
\end{tabular*}
\end{table*}

\subsection{Ablation Study}\label{Ablation}
\label{sec:ablation}
The ablation study compares factorization, bi-sparse optimization, and channel-scoring
variants on the CIFAR-10 dataset, then examines their scores and retained-channel sets
to explain the resulting selections.

\subsubsection{Component Contributions}

Table \ref{tab:component_ablation} reports the ResNet-56 ablation results on the CIFAR-10 dataset, where accuracy is reported as
the mean and standard deviation over three random seeds. It can be concluded that
\begin{itemize}
    \item Without pruning, BSMF achieves 93.09\% accuracy versus 93.26\% for the dense baseline, with unchanged FLOPs and 0.85M parameters. The 0.17-percentage-point decrease shows that factorization, bi-sparse optimization, and filter recovery preserve near-baseline performance before channel removal.

    \item With identical $A$-row scoring, 40\% FLOPs reduction, and 0.44M parameters, adding bi-sparse optimization to factorization-only pruning raises mean accuracy from 93.83\% to 93.99\% and reduces the observed standard deviation from 0.18 to 0.07, supporting its effectiveness beyond factorization alone.

    \item At 40\% FLOPs reduction, adding factorization and bi-sparse optimization to direct $\ell_2$ pruning raises mean accuracy from 93.65\% to 93.74\% and reduces parameters from 0.55M to 0.52M. This modest benefit with recovered-filter $\ell_2$ scoring supports the pipeline's effectiveness and provides a reference for evaluating $A$-row scoring.

    \item Among the three BSMF scoring variants, filter-$\ell_1$ improves accuracy over filter-$\ell_2$ from 93.74\% to 93.91\%, while $A$-row scoring achieves the highest accuracy (93.99\%) with the fewest parameters (0.44M). By scoring the regularized coefficient groups rather than recovered filters, $A$-row scoring directly connects sparse optimization to channel selection, yielding the best pruning performance among the evaluated configurations.
\end{itemize}

Overall, BSMF maintains near-baseline accuracy before pruning, benefits from bi-sparse optimization, and improves conventional norm-based pruning. Notably, with A-row scoring, it achieves the highest accuracy with the fewest parameters among the evaluated configurations.

\begin{figure*}[t]
\centering

\subfigure[Baseline]{
\includegraphics[width=0.30\textwidth]{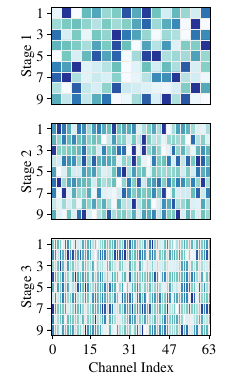}}
\hfill
\subfigure[BSMF $A$-row]{
\includegraphics[width=0.30\textwidth]{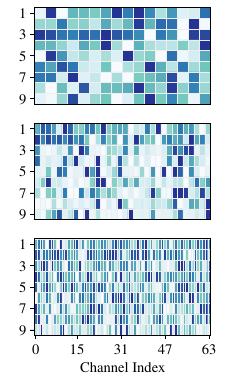}}
\hfill
\subfigure[Final Mask]{
\includegraphics[width=0.30\textwidth]{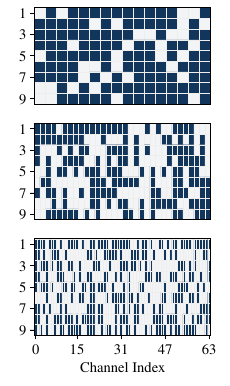}}
\caption{Channel-selection heatmaps for ResNet-56, where (a)-(c) show baseline filter $\ell_2$-norms, BSMF $A$-row scores, and retention masks, respectively.}
\label{fig:channel_selection_heatmap}
\end{figure*}
\begin{figure*}[tp]
\centering
\includegraphics[width=0.98\textwidth]{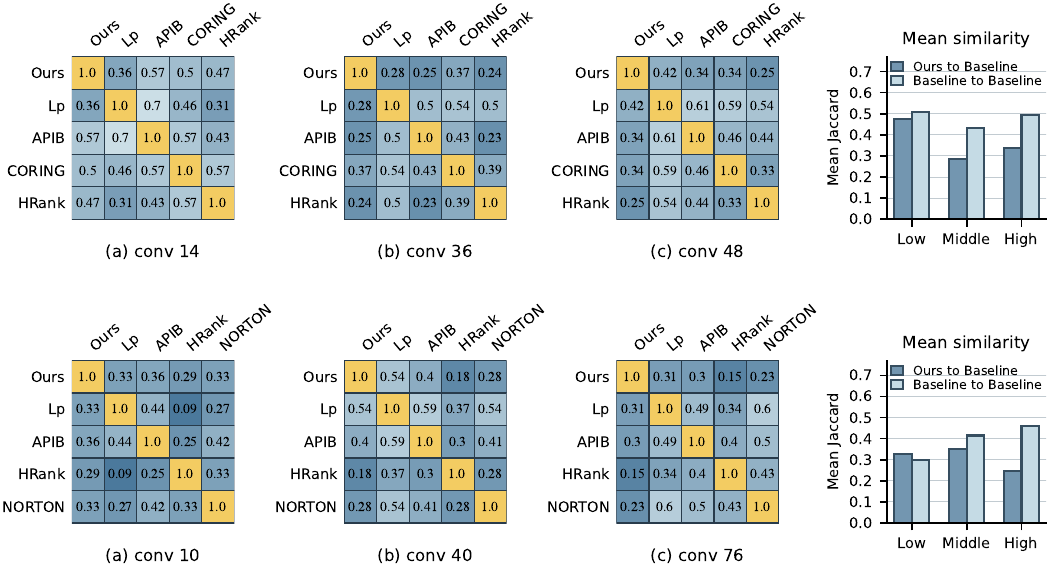}
\caption{Pairwise Jaccard similarity of retained-channel sets on the CIFAR-10 dataset, where the top and bottom rows show ResNet-56 and ResNet-110, respectively.}
\label{fig:jaccard_similarity}
\end{figure*}

\subsubsection{Effect of Factorized Channel Scoring}
Fig. \ref{fig:channel_selection_heatmap} compares baseline filter $\ell_2$-norms (a), optimized $A$-row scores (b), and the final retention mask (c) for ResNet-56.
The scores in (b), defined in Eq. \eqref{eq:pipeline_channel_score}, measure each channel's coefficient strength in a shared basis. Compared with (a), they show clearer contrast among channels, particularly in the deeper stages, consistent with row-wise regularization suppressing weak coefficient groups. The mask in (c) is constructed from these scores under the network's structural constraints, linking the regularized factorized representation to channel selection.

\subsubsection{Comparison of Retained-Channel Sets}

To examine how its channel selections differ from those of existing criteria, we compare the retained-channel sets at the same layer using the Jaccard similarity. For
layer $i$, it is defined as
\begin{equation}
J_i(A,B)
=
\frac{\left|S_i^{(A)}\cap S_i^{(B)}\right|}
{\left|S_i^{(A)}\cup S_i^{(B)}\right|},
\label{eq:jaccard_similarity}
\end{equation}
where $S_i^{(A)}$ and $S_i^{(B)}$ are the channel sets retained by methods $A$ and $B$, respectively.

Fig. \ref{fig:jaccard_similarity} shows pairwise similarities for representative layers of ResNet-56 and ResNet-110. The mean-similarity plots indicate that BSMF's selections are generally less similar to those of the other criteria in the middle and deeper layers, while the trend in shallow layers varies between the two networks. This suggests that BSMF can retain
channel sets that differ from those selected by conventional criteria, consistent with its factorized channel-scoring mechanism. However, the reasons for the layer-dependent differences in similarity remain unclear
and warrant further investigation.

The ablations show higher mean accuracy with bi-sparse optimization than with factorization alone at the tested budget, while the retained-set comparisons reveal selection differences from the evaluated criteria. BSMF's key feature is the use of regularized, channel-aligned coefficient groups in a learned shared basis for importance scoring, directly linking sparse optimization to channel selection.

\subsection{Discussion}\label{Discussion}
\label{sec:discussion}
This section provides a convergence analysis, examines how regularization orders and architectural constraints affect pruning, and uses effective weight distributions to relate factor sparsity to channel selection.

\begin{figure}[t]
\centering

\begin{minipage}[t]{0.35\linewidth}
  \centering
  \subfigure[CIFAR-10 on VGG-16]{%
    \includegraphics[width=\linewidth]{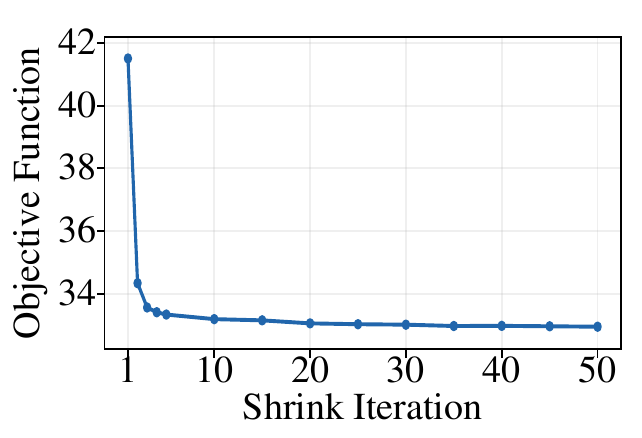}}
\end{minipage}\qquad
\begin{minipage}[t]{0.35\linewidth}
  \centering
  \subfigure[CIFAR-10 on ResNet-56]{%
    \includegraphics[width=\linewidth]{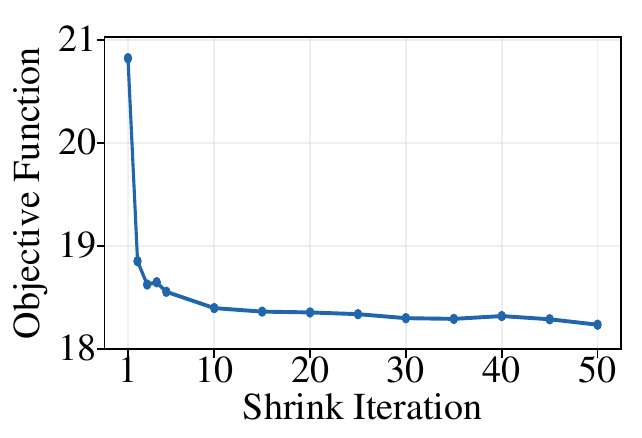}}
\end{minipage}

\par\vspace{0.8em}

\begin{minipage}[t]{0.35\linewidth}
  \centering
  \subfigure[CIFAR-10 on ResNet-110]{%
    \includegraphics[width=\linewidth]{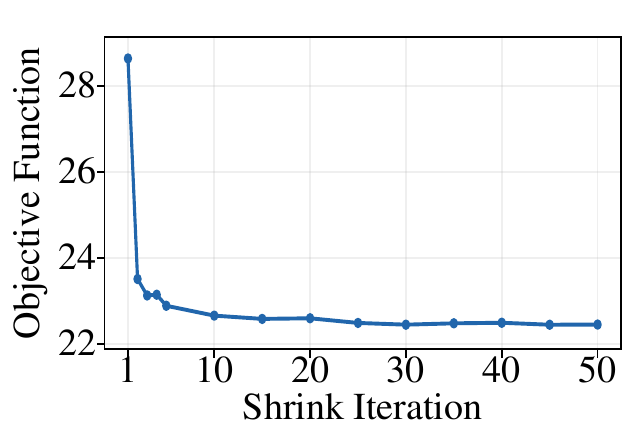}}
\end{minipage}\qquad
\begin{minipage}[t]{0.35\linewidth}
  \centering
  \subfigure[ImageNet on ResNet-50]{%
    \includegraphics[width=\linewidth]{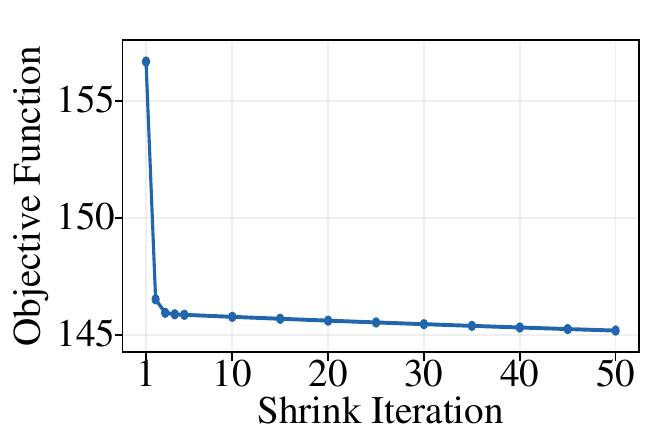}}
\end{minipage}

\caption{Convergence curves of BSMF on different models and
datasets, where (a)-(d) show the objective function values.}
\label{fig:numerical_stabilization}
\end{figure}

\subsubsection{Convergence Analysis}
Fig. \ref{fig:numerical_stabilization}(a)-(d) shows the objective values of problem \eqref{eq:main_obj} during Algorithm \ref{alg:fgsp_admm} for VGG-16, ResNet-56, and ResNet-110 on the CIFAR-10 dataset, and ResNet-50 on the ImageNet-1K dataset, respectively.
Across all four settings, the objective decreases rapidly in the first few shrink iterations and then approaches a stable level, with only small subsequent changes.
BSMF couples the row- and column-group penalties in Eq. \eqref{eq:main_obj} to regularize channel coefficients and the shared basis, respectively.
The observed stabilization provides numerical support for optimizing this coupled nonconvex formulation through alternating ADMM.

\subsubsection{Sensitivity to Regularization Orders}
\label{sec:exponent_sensitivity}
\begin{figure*}[t]
\centering
\subfigure[ResNet-56]{\includegraphics[width=0.44\textwidth]{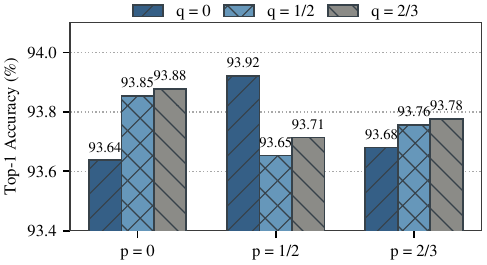}}\hspace{0.04\textwidth}%
\subfigure[ResNet-110]{\includegraphics[width=0.44\textwidth]{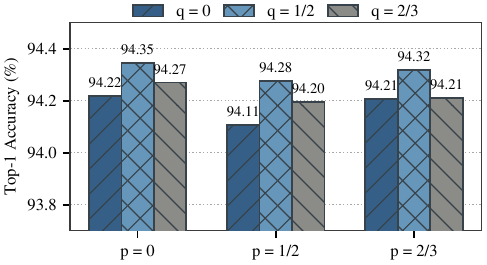}}
\caption{Sensitivity to $p$ and $q$ on the CIFAR-10 dataset, where (a,b) show mean best-checkpoint Top-1 accuracy over three seeds for ResNet-56 and ResNet-110, respectively.}
\label{fig:pq_sensitivity}
\end{figure*}

We evaluate nine pairs $(p,q)\in\{0,1/2,2/3\}^{2}$ for the regularization orders in Eq. \eqref{eq:main_obj} on ResNet-56 and ResNet-110 using the CIFAR-10 dataset.
Fig. \ref{fig:pq_sensitivity} reports the mean Top-1 accuracy over three random seeds.
The results suggest selecting $p$ and $q$ jointly. For ResNet-110, $q=1/2$ generally performs well, suggesting half-order regularization on the shared-basis side while considering several candidate orders on the coefficient side.
Here, $p$ and $q$ govern row sparsity in channel coefficients and column sparsity in the shared basis, respectively.
Overall, the mean accuracy ranges span 0.28 and 0.24 percentage points for ResNet-56 and ResNet-110, respectively, suggesting relative insensitivity to these orders within the tested range.
However, the limited grid and small differences between configurations require further evaluation with more seeds and combined consideration of regularization weights, factorization rank, and fine-tuning settings to determine preferred choices.
Their applicability to other models also requires validation.

\subsubsection{Channel Allocation and Deployment}
\label{sec:channel_allocation}
\begin{figure*}[t]
\centering

\subfigure[VGG-16]{
\includegraphics[width=0.46\textwidth]{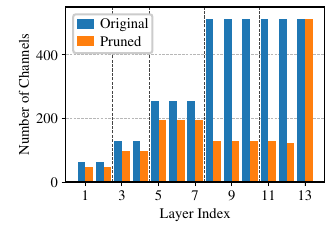}}
\hfill
\subfigure[ResNet-56]{
\includegraphics[width=0.46\textwidth]{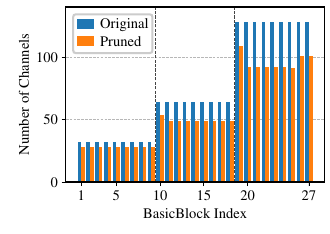}}
\\
\subfigure[ResNet-110]{
\includegraphics[width=0.46\textwidth]{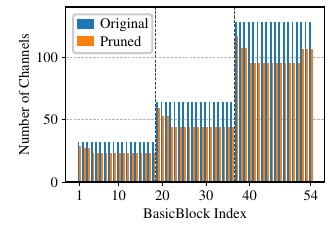}}
\hfill
\subfigure[ResNet-50]{
\includegraphics[width=0.46\textwidth]{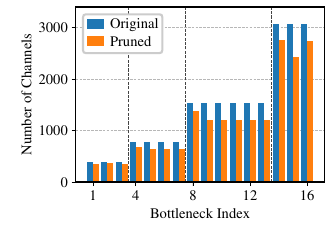}}
\caption{Channel counts before and after pruning, where counts are shown per VGG-16 layer and summed over the main-branch convolutions of each ResNet block.}
\label{fig:pruned_channel_profiles}
\end{figure*}

Fig. \ref{fig:pruned_channel_profiles} illustrates channel protection by layer function and pruning sensitivity \cite{ye2023pagcp}. VGG-16 (a) compresses earlier layers while preserving its final output width for the classifier, consistent with ThiNet-Tiny's milder final-layer pruning \cite{luo2019thinet}. ResNet-56 and 110 (b,c) retain block output widths for shortcut compatibility and protect the stem and Stage 1 against early information loss, with stronger protection at the entrances of Stages 2 and 3 for cross-scale transfer and at the end of Stage 3 for classifier inputs. ResNet-50 (d) prunes bottleneck interiors while retaining block outputs and projection shortcuts, as in ThiNet \cite{luo2019thinet}. These constraints preserve feature transfer, residual addition, and classification while concentrating compression on internal computation. Within these constraints, BSMF selects channels using coefficient-row scores learned through bi-sparse shared filter modeling, then recombines the factors and physically removes channels to obtain a compact network with standard dense convolutions.

\subsubsection{Effective Weight Distributions}
\begin{figure*}[t]
\centering

\subfigure[ResNet-56: original]{
\begingroup%
\setlength{\unitlength}{0.002712448008\textwidth}%
\begin{picture}(84.794252,92)%
\put(0,0){\includegraphics[width=0.23\textwidth]{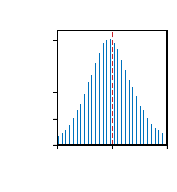}}%
\footnotesize\rmfamily%
\put(27.613151,14.709129){\makebox(0,0){$-0.1$}}%
\put(53.884319,14.709129){\makebox(0,0){0}}%
\put(80.152924,14.709129){\makebox(0,0){0.1}}%
\put(53.870886,4.706001){\makebox(0,0){Weight value}}%
\put(23.815714,21.095055){\makebox(0,0)[r]{0}}%
\put(23.815714,33.613243){\makebox(0,0)[r]{0.5}}%
\put(23.815714,46.131432){\makebox(0,0)[r]{1}}%
\put(23.815714,71.167795){\makebox(0,0)[r]{2}}%
\put(6.953464,49.940065){\makebox(0,0){\rotatebox[origin=c]{90}{Frequency}}}%
\put(27.615713,85.009760){\makebox(0,0)[l]{\texttimes10\textsuperscript{4}}}%
\end{picture}%
\endgroup%
}
\hfill
\subfigure[ResNet-110: original]{
\begingroup%
\setlength{\unitlength}{0.002712448008\textwidth}%
\begin{picture}(84.794252,92)%
\put(0,0){\includegraphics[width=0.23\textwidth]{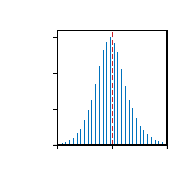}}%
\footnotesize\rmfamily%
\put(27.613152,14.709129){\makebox(0,0){$-0.1$}}%
\put(53.884315,14.709129){\makebox(0,0){0}}%
\put(80.152931,14.709129){\makebox(0,0){0.1}}%
\put(53.870880,4.706001){\makebox(0,0){Weight value}}%
\put(23.815712,21.095055){\makebox(0,0)[r]{0}}%
\put(23.815712,38.307564){\makebox(0,0)[r]{2}}%
\put(23.815712,55.520058){\makebox(0,0)[r]{4}}%
\put(23.815712,72.732569){\makebox(0,0)[r]{6}}%
\put(12.203461,49.940065){\makebox(0,0){\rotatebox[origin=c]{90}{Frequency}}}%
\put(27.615715,85.009760){\makebox(0,0)[l]{\texttimes10\textsuperscript{4}}}%
\end{picture}%
\endgroup%
}
\hfill
\subfigure[ResNet-50: original]{
\begingroup%
\setlength{\unitlength}{0.002712448008\textwidth}%
\begin{picture}(84.794252,92)%
\put(0,0){\includegraphics[width=0.23\textwidth]{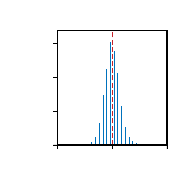}}%
\footnotesize\rmfamily%
\put(27.613144,14.709129){\makebox(0,0){$-0.1$}}%
\put(53.884323,14.709129){\makebox(0,0){0}}%
\put(80.152924,14.709129){\makebox(0,0){0.1}}%
\put(53.870872,4.706001){\makebox(0,0){Weight value}}%
\put(23.815704,21.095055){\makebox(0,0)[r]{0}}%
\put(23.815704,37.295067){\makebox(0,0)[r]{0.5}}%
\put(23.815704,53.495064){\makebox(0,0)[r]{1}}%
\put(23.815704,69.695063){\makebox(0,0)[r]{1.5}}%
\put(6.953461,49.940065){\makebox(0,0){\rotatebox[origin=c]{90}{Frequency}}}%
\put(27.615707,84.976948){\makebox(0,0)[l]{\texttimes10\textsuperscript{6}}}%
\end{picture}%
\endgroup%
}
\hfill
\subfigure[VGG-16: original]{
\begingroup%
\setlength{\unitlength}{0.002712448008\textwidth}%
\begin{picture}(84.794252,92)%
\put(0,0){\includegraphics[width=0.23\textwidth]{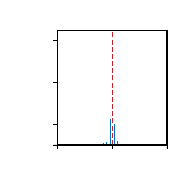}}%
\footnotesize\rmfamily%
\put(27.613136,14.709129){\makebox(0,0){$-0.1$}}%
\put(53.884315,14.709129){\makebox(0,0){0}}%
\put(80.152931,14.709129){\makebox(0,0){0.1}}%
\put(53.870888,4.706001){\makebox(0,0){Weight value}}%
\put(23.815712,21.095055){\makebox(0,0)[r]{0}}%
\put(23.815712,31.109612){\makebox(0,0)[r]{1}}%
\put(23.815712,51.138695){\makebox(0,0)[r]{3}}%
\put(23.815712,71.167795){\makebox(0,0)[r]{5}}%
\put(12.203438,49.940065){\makebox(0,0){\rotatebox[origin=c]{90}{Frequency}}}%
\put(27.615700,84.976948){\makebox(0,0)[l]{\texttimes10\textsuperscript{6}}}%
\end{picture}%
\endgroup%
}
\\
\subfigure[ResNet-56: effective]{
\begingroup%
\setlength{\unitlength}{0.002712448008\textwidth}%
\begin{picture}(84.794252,92)%
\put(0,0){\includegraphics[width=0.23\textwidth]{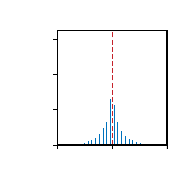}}%
\footnotesize\rmfamily%
\put(27.613151,14.709129){\makebox(0,0){$-0.1$}}%
\put(53.884319,14.709129){\makebox(0,0){0}}%
\put(80.152924,14.709129){\makebox(0,0){0.1}}%
\put(53.870886,4.706001){\makebox(0,0){Weight value}}%
\put(23.815714,21.095070){\makebox(0,0)[r]{0}}%
\put(23.815714,38.042762){\makebox(0,0)[r]{0.4}}%
\put(23.815714,54.990455){\makebox(0,0)[r]{0.8}}%
\put(23.815714,71.938144){\makebox(0,0)[r]{1.2}}%
\put(6.953464,49.940056){\makebox(0,0){\rotatebox[origin=c]{90}{Frequency}}}%
\put(27.615713,84.955074){\makebox(0,0)[l]{\texttimes10\textsuperscript{5}}}%
\end{picture}%
\endgroup%
}
\hfill
\subfigure[ResNet-110: effective]{
\begingroup%
\setlength{\unitlength}{0.002712448008\textwidth}%
\begin{picture}(84.794252,92)%
\put(0,0){\includegraphics[width=0.23\textwidth]{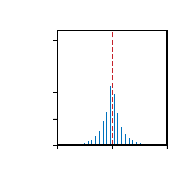}}%
\footnotesize\rmfamily%
\put(27.613152,14.709129){\makebox(0,0){$-0.1$}}%
\put(53.884315,14.709129){\makebox(0,0){0}}%
\put(80.152931,14.709129){\makebox(0,0){0.1}}%
\put(53.870880,4.706001){\makebox(0,0){Weight value}}%
\put(23.815712,21.095070){\makebox(0,0)[r]{0}}%
\put(23.815712,33.613258){\makebox(0,0)[r]{0.5}}%
\put(23.815712,46.131431){\makebox(0,0)[r]{1}}%
\put(23.815712,71.167789){\makebox(0,0)[r]{2}}%
\put(6.953461,49.940056){\makebox(0,0){\rotatebox[origin=c]{90}{Frequency}}}%
\put(27.615715,84.955074){\makebox(0,0)[l]{\texttimes10\textsuperscript{5}}}%
\end{picture}%
\endgroup%
}
\hfill
\subfigure[ResNet-50: effective]{
\begingroup%
\setlength{\unitlength}{0.002712448008\textwidth}%
\begin{picture}(84.794252,92)%
\put(0,0){\includegraphics[width=0.23\textwidth]{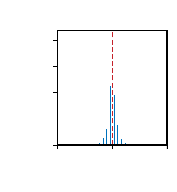}}%
\footnotesize\rmfamily%
\put(27.613144,14.709129){\makebox(0,0){$-0.1$}}%
\put(53.884323,14.709129){\makebox(0,0){0}}%
\put(80.152924,14.709129){\makebox(0,0){0.1}}%
\put(53.870872,4.706001){\makebox(0,0){Weight value}}%
\put(23.815704,21.095070){\makebox(0,0)[r]{0}}%
\put(23.815704,46.131431){\makebox(0,0)[r]{2}}%
\put(23.815704,58.649604){\makebox(0,0)[r]{3}}%
\put(23.815704,71.167789){\makebox(0,0)[r]{4}}%
\put(12.203461,49.940056){\makebox(0,0){\rotatebox[origin=c]{90}{Frequency}}}%
\put(27.615707,84.976947){\makebox(0,0)[l]{\texttimes10\textsuperscript{6}}}%
\end{picture}%
\endgroup%
}
\hfill
\subfigure[VGG-16: effective]{
\begingroup%
\setlength{\unitlength}{0.002712448008\textwidth}%
\begin{picture}(84.794252,92)%
\put(0,0){\includegraphics[width=0.23\textwidth]{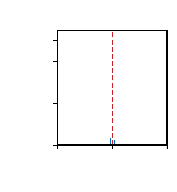}}%
\footnotesize\rmfamily%
\put(27.613136,14.709129){\makebox(0,0){$-0.1$}}%
\put(53.884315,14.709129){\makebox(0,0){0}}%
\put(80.152931,14.709129){\makebox(0,0){0.1}}%
\put(53.870888,4.706001){\makebox(0,0){Weight value}}%
\put(23.815712,21.095070){\makebox(0,0)[r]{0}}%
\put(23.815712,41.124153){\makebox(0,0)[r]{0.4}}%
\put(23.815712,61.153251){\makebox(0,0)[r]{0.8}}%
\put(23.815712,71.167789){\makebox(0,0)[r]{1}}%
\put(6.953438,49.940056){\makebox(0,0){\rotatebox[origin=c]{90}{Frequency}}}%
\put(27.615700,85.075390){\makebox(0,0)[l]{\texttimes10\textsuperscript{7}}}%
\end{picture}%
\endgroup%
}
\caption{Weight distributions, where (a)-(d) show original weights and (e)-(h) show effective weights constructed using normalized $A$-row scores.}
\label{fig:weight_distribution}
\end{figure*}

Fig. \ref{fig:weight_distribution} compares original weights (a)-(d) with effective weights (e)-(h), defined for visualization as $W_{\mathrm{eff},l}=\operatorname{diag}(g_l)W_{\mathrm{rec},l}$, where $W_{\mathrm{rec},l}$ contains dense filters after factor reconstruction and recovery, and $g_{l,i}=0.05+0.95\widetilde{s}_{l,i}$ uses the normalized $A$-row score in Eq. \eqref{eq:pipeline_score_normalization}. Across all four architectures, effective weights concentrate more strongly near zero, as illustrated in (a,e). Since each $A$ row represents an output channel, score-based attenuation visualizes the learned coefficient-row sparsity. Under structural constraints, BSMF converts these scores into pruning masks, translating the sparse structure learned through bi-sparse shared filter modeling into a channel-pruning strategy.

\section{Conclusion}\label{sec:conclusion}

This paper proposes a novel structured channel pruning method called BSMF. By expressing convolutional filters as channel-aligned coefficients over a shared basis and applying structured sparsity regularization terms to both factors, our BSMF assesses channel importance within a representation that captures relationships among filters. In  algorithms, an efficient alternating minimization strategy is designed based on the ADMM framework. Experiments on the CIFAR-10 and ImageNet-1K datasets demonstrate competitive classification accuracy with reduced FLOPs and parameter counts. These results confirm the significant potential of our proposed pruning model.

Of course, there remain some limitations. On one hand, the compression stage requires a significant number of iterative computations, thus a faster and more efficient algorithm should be developed. On the other hand, this work focuses exclusively on CNN-based image models and extending it to large language models would also be a meaningful direction.

\section*{Acknowledgments}
This work was supported in part by the National Natural Science Foundation of China under Grants 62204044 and 12371306, the State Key Laboratory of Integrated Chips and Systems under Grant SKLICS-K202302, and the Shenzhen Nature Science Research Foundation Key Project under Grant ZDYJ20251211121303005.

\def\DOIprefix{}
\def\doi#1{}
\biboptions{sort&compress}
\IfFileExists{mybibfile.bib}{%
  \bibliographystyle{elsarticle-num-names}
  \bibliography{mybibfile}
}{%
  \bibliographystyle{\BSMFProjectRoot/elsarticle-num-names}
  \bibliography{\BSMFProjectRoot/mybibfile}
}

\end{document}